\documentclass[runningheads]{llncs}

\usepackage[T1]{fontenc}
\usepackage{orcidlink}
\hypersetup{hidelinks}

\usepackage{graphicx}
\usepackage{booktabs}
\usepackage{multirow}
\usepackage{array}
\usepackage{amsmath}
\usepackage{amssymb}
\usepackage[table]{xcolor}
\usepackage{colortbl}
\usepackage{enumitem}
\usepackage{algorithm}
\usepackage{algpseudocode}
\usepackage{amssymb}
\usepackage{algorithm}
\usepackage{algpseudocode}
\usepackage{xcolor}

\usepackage{marvosym}

\usepackage{float}

\begin{document}
\title{RouteGraph-Mona: Confusion-Aware Routing Fine-Tuning for Mineral Image Classification}
\titlerunning{RouteGraph-Mona}
%
%
%
\author{
  Jierui Li$^{\star}$\inst{1}\orcidlink{0009-0007-3540-2279},
  Zhiyuan Qi\thanks{Equal contribution.}\inst{2}\orcidlink{0009-0001-8713-0312},
  Hao Zhu\inst{1}\orcidlink{0000-0003-4411-0933},
  Yufan Liu\inst{3}\orcidlink{0009-0007-0390-9219},
  Jixian Liu\inst{1}\orcidlink{0009-0006-3295-6404},
  Shaojie Jiang\inst{1}\orcidlink{0009-0004-3703-720X},
  Jianda Wang\inst{1}\orcidlink{0009-0006-5497-6537},
  Yaqi Liu\inst{3}\orcidlink{0009-0004-3867-2950},
  Xiaotong Li\textsuperscript{\Letter}\inst{1}\orcidlink{0000-0001-9960-5373},
  Wei Wang\textsuperscript{\Letter}\inst{3}\orcidlink{0009-0002-5769-5870}
}

\authorrunning{J. Li, Z. Qi et al.}

\institute{
  Xidian University, Xi'an, China \and
  Tsinghua University, Beijing, China \and
  Guangdong Laboratory of Artificial Intelligence and Digital Economy (SZ), China
    \email{jerryli@stu.xidian.edu.cn, qzy24@mails.tsinghua.edu.cn}\\
  \email{lixiaotong@xidian.edu.cn, wangwei@gml.ac.cn}\\
}

\maketitle
\begingroup
\renewcommand{\thefootnote}{\Letter}
\footnotetext{Corresponding authors.}
\endgroup
\begin{abstract}
Mineral image classification is important for geological exploration and resource development, but it remains challenging due to substantial intra-class variations in appearance and high inter-class visual similarity.
\textbf{M}ulti-c\textbf{o}g\textbf{n}itive Visual \textbf{A}dapter (\textbf{Mona}) is a vision-oriented parameter-efficient adapter that adapts pre-trained visual models by tuning only a few parameters.
However, Mona statically aggregates responses from multiple scales, limiting its ability to accommodate sample-specific scale preferences and model confusion among visually similar mineral categories.
To address this issue, we propose \textbf{RouteGraph-Mona}, a lightweight route-space regularization method built on Mona.
Specifically, we replace Mona's static multi-scale aggregation with sample-adaptive routing. The resulting branch-selection behavior defines a compact routing space that captures each image's scale preferences.
We then regularize the resulting routing signatures with class-wise route anchors and confusion-weighted margins. The route anchors encourage class-consistent routing patterns, while the margins promote greater separation between visually similar categories in the routing space. Experiments on three public mineral image datasets with two visual backbones show that RouteGraph-Mona consistently outperforms Mona in mean accuracy and remains competitive with representative fine-tuning methods and mineral image classification baselines. The project repository is available at \url{https://github.com/rgmona/RouteGraph-Mona}.

\begin{figure}[t]
\centering
\includegraphics[width=0.90\linewidth]{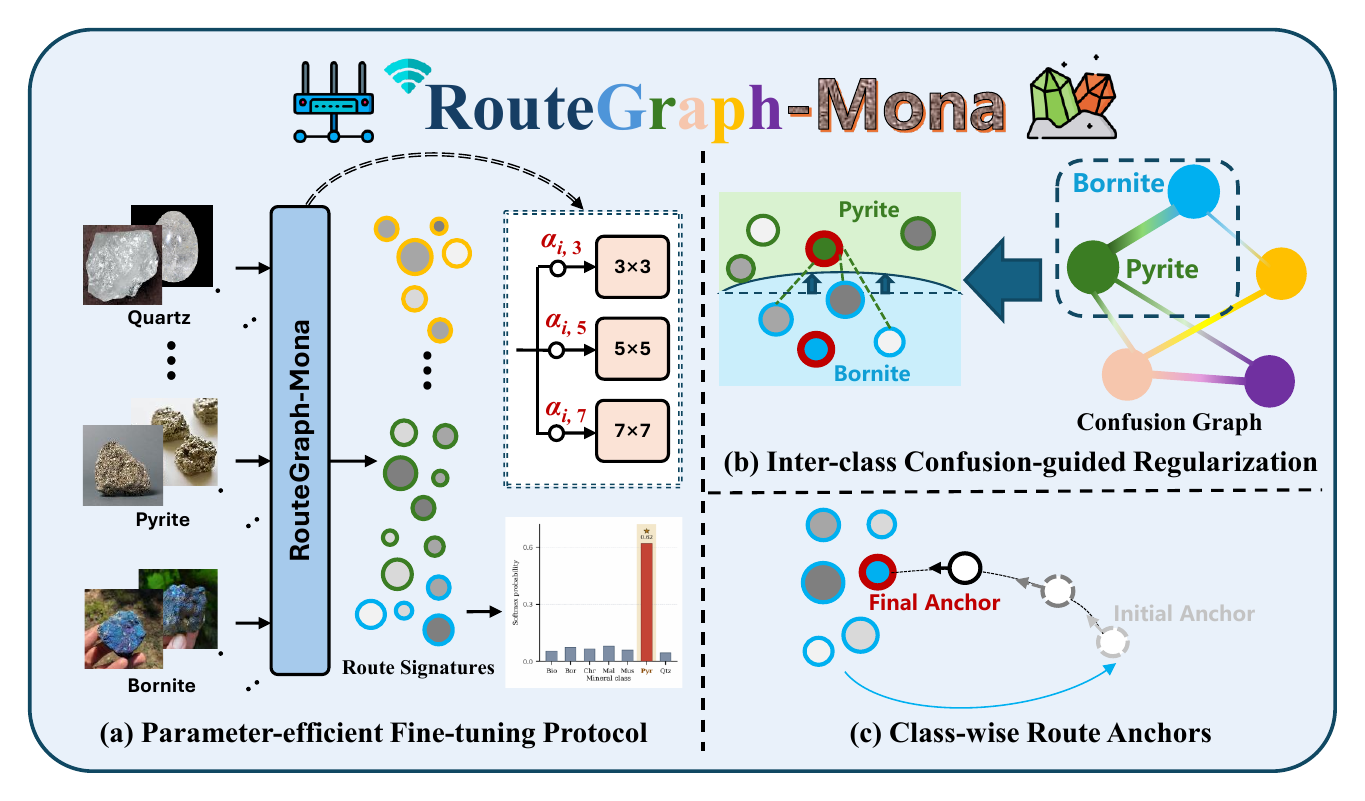}
\caption{Overview of RouteGraph-Mona. The method summarizes adaptive route weights over Mona's multi-scale branches into route signatures and regularizes them with class-wise route anchors and an online confusion graph for confusion-aware route-space separation.}
\label{fig:tldr1}
\end{figure}


\end{abstract}
\section{Introduction}
\label{intro}

Mineral image classification is important for geological exploration, resource development, and resource evaluation~\cite{lou2020review}.
Traditional mineral identification relies on expert knowledge and specialized instruments, including Raman spectroscopy, X-ray fluorescence spectroscopy, X-ray diffraction, scanning electron microscopy, and automated mineralogy systems~\cite{haskin1997raman,ali2022xrd,ali2023sem}.
Although reliable, these methods are costly and complicated, limiting large-scale, low-cost automated identification~\cite{zeng2021mineral,jia2021mineralphotos,wu2022multilabel}.
In contrast, image-based classification is easy to deploy, and existing CNN-based, feature-fusion-based, DenseNet-based, YOLOv8-based, and Transformer-based methods have shown strong potential for mineral recognition~\cite{asiedu2021minet,jia2021mineralphotos,wu2022multilabel,swinmin2024,long2025yolov8cls,attallah2025mineral}.

However, deep classification models require large annotated datasets and substantial computational resources, whereas collecting and labeling mineral images is difficult~\cite{asiedu2021minet,jia2021mineralphotos,wu2022multilabel,swinmin2024}. 
Pre-training and fine-tuning address this challenge by transferring visual knowledge from large-scale models to downstream tasks~\cite{yosinski2014transferable,kornblith2019better,kolesnikov2020big}. 
Full fine-tuning adapts all model parameters, while parameter-efficient fine-tuning updates only lightweight modules and freezes most of the backbone~\cite{houlsby2019parameter,hu2022lora,jia2022visual,he2023parameter}.

Among parameter-efficient visual fine-tuning methods, Mona is a representative vision-oriented adapter~\cite{yin2025}. 
It introduces filtering structures and scaled normalization to improve visual transferability~\cite{yin2025}. 
However, its direct application to mineral image classification remains challenging. 
Mineral images often exhibit substantial intra-class variations in appearance and high inter-class visual similarity, where samples from the same category may vary in shape, color, and texture, while different categories may share similar local cues~\cite{jia2021mineralphotos,wu2022multilabel,swinmin2024}. 
Under these conditions, Mona's static averaging of multi-scale branch responses may fail to capture image-specific scale preferences, while standard classification-driven adaptation does not explicitly exploit category-level confusion.

We observe that routing weights induced by adaptive multi-scale aggregation define a low-dimensional route space that reflects image-specific preferences over different receptive-field scales. 
Compared with visual feature embeddings, this space emphasizes branch-selection patterns rather than appearance details, making it suitable for mineral images with substantial intra-class variation. 
We therefore regularize the routing signatures generated by adaptive routing to promote class-consistent routing patterns and improve separation between visually confusing categories.

To this end, we propose \textbf{RouteGraph-Mona}, a lightweight route-space regularization method for mineral image classification. 
As illustrated in Fig.~\ref{fig:tldr1}, RouteGraph-Mona replaces Mona's static multi-scale aggregation with sample-adaptive routing, constructs route signatures from the routing weights, and regularizes them with class-wise route anchors and a confusion-weighted route margin. 
The route anchors provide category-level route prototypes, while the online
confusion graph assigns larger weights to route-margin penalties for visually
confusing category pairs. 
This design improves relative route-space separation without requiring labels, graph updates, or auxiliary losses during inference.

The main contributions are summarized as follows:
\begin{itemize}
\item We propose \textbf{RouteGraph-Mona}, which introduces sample-adaptive routing into Mona's multi-scale branches and forms a low-dimensional route space for mineral image classification.

\item We design class-wise route anchors and an online confusion graph to regularize routing signatures, guiding category-level routing tendencies and enlarging route-level separation between visually ambiguous categories.

\item Experiments on three public mineral image datasets and two visual backbones show that RouteGraph-Mona improves the mean accuracy over Mona across the evaluated settings and achieves competitive performance against representative baselines.
\end{itemize}
\section{Related Work}
\label{related}

\subsection{Mineral Image Classification Methods}

Early mineral image classification methods mainly relied on handcrafted descriptors, such as color, texture, and shape features, together with traditional machine learning classifiers \cite{lou2020review,zeng2021mineral,jia2021mineralphotos}. With the development of deep learning, CNN-based and Transformer-based models have been increasingly used for mineral image recognition, enabling discriminative visual representations to be learned directly from images \cite{asiedu2021minet,jia2021mineralphotos,wu2022multilabel,swinmin2024}. Recent studies have further explored DenseNet variants, YOLOv8-CLS, and task-specific transformer models for mineral classification, typically combined with transfer learning, feature fusion, data augmentation, or multi-scale contextual modeling to improve recognition performance \cite{attallah2025mineral,long2025yolov8cls,swinmin2024}. These methods mainly improve mineral recognition by designing stronger classifiers or feature extractors. In contrast, our work focuses on parameter-efficient adaptation of pre-trained visual backbones, while regularizing class-aware routing behavior in the route space.

\subsection{Parameter-Efficient Visual Fine-Tuning}

Pre-training and fine-tuning have become a common paradigm for transferring large-scale visual models to downstream recognition tasks \cite{yosinski2014transferable,kornblith2019better,kolesnikov2020big}. Although full fine-tuning usually provides strong adaptation, it updates all model parameters and incurs high computational and storage costs. Parameter-efficient fine-tuning methods reduce this cost by updating only a small number of parameters or lightweight modules, such as adapters, LoRA, visual prompt tuning, and other efficient adaptation modules \cite{houlsby2019parameter,hu2022lora,jia2022visual,he2023parameter}. Mona further introduces vision-friendly filtering structures and scaled normalization, providing an effective adapter design for visual transfer tasks \cite{yin2025}. However, these methods are mainly optimized by the classification objective and do not explicitly model the category-level structure of mineral images. To address this limitation, RouteGraph-Mona regularizes the route space induced by adaptive multi-scale aggregation, using class-wise route anchors and an online confusion graph to improve route-level separation for visually ambiguous mineral categories.

\section{Method}

\subsection{Overall Framework}

\begin{figure}[t]
\centering
\includegraphics[width=0.83\linewidth]{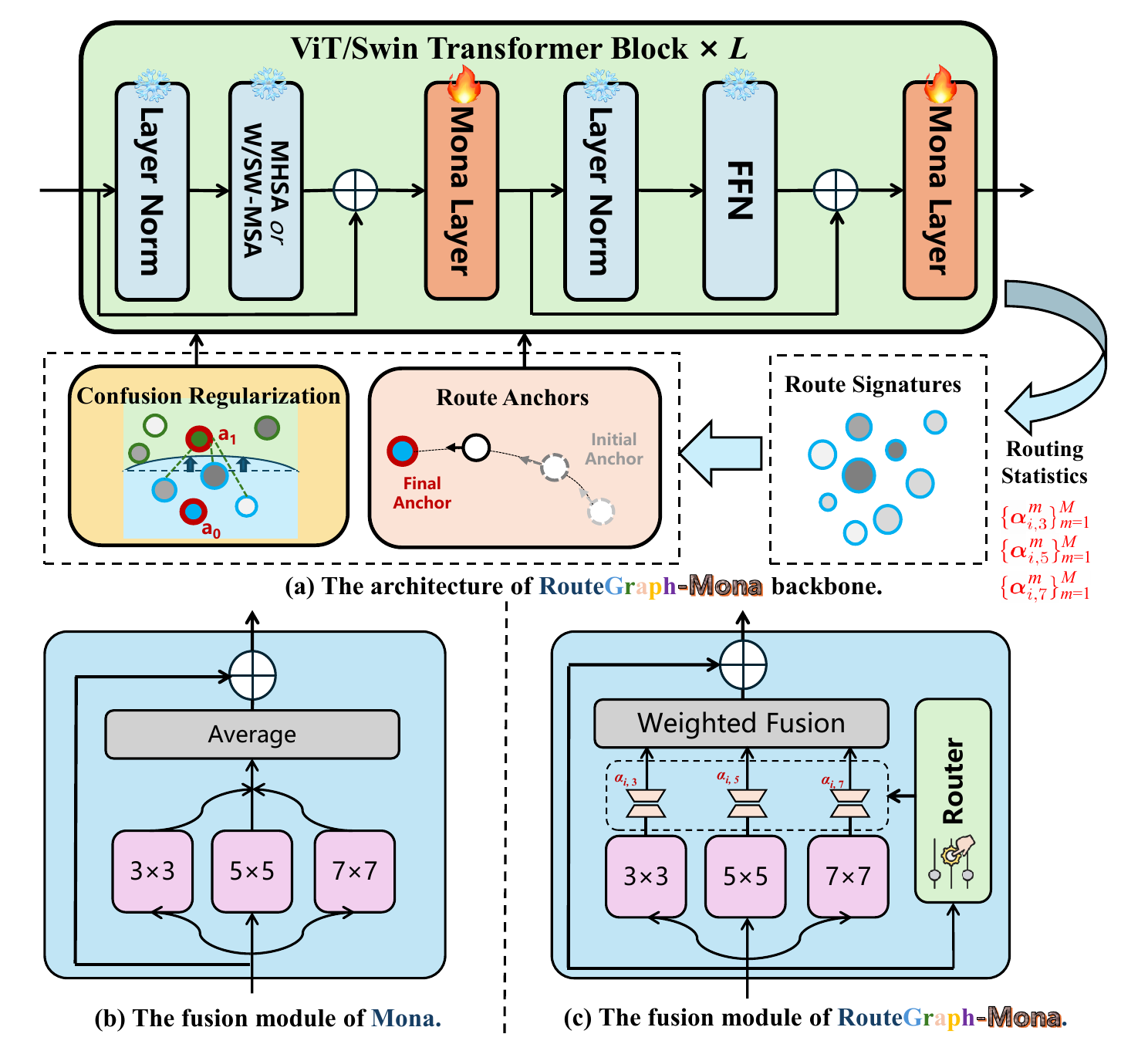}
\caption{Architecture of RouteGraph-Mona.
(a) Overall backbone with route signatures and route-space regularization.
(b) Static average fusion in Mona.
(c) Router-based weighted fusion in RouteGraph-Mona; anchors and the confusion graph regularize the route space.}
\label{fig:tldr2}
\end{figure}

We propose RouteGraph-Mona, a lightweight route-space regularization method built on Mona and designed for mineral image classification.
As shown in Fig.~\ref{fig:tldr2}(a), RouteGraph-Mona replaces Mona's static averaging of multi-scale branch responses with sample-adaptive routing and summarizes the resulting routing weights into compact route signatures.
To avoid directly constraining diverse visual appearances, these signatures describe Mona's branch-selection behavior rather than visual feature embeddings.
They are further regularized by class-wise route anchors and an online confusion graph, so that visually confusing categories can be better separated in the route space.
During inference, RouteGraph-Mona uses a standard forward pass and predicts from the classification logits.

\subsection{Parameter-efficient Fine-tuning Protocol}

RouteGraph-Mona follows a parameter-efficient fine-tuning protocol.
To adapt pretrained visual backbones to mineral recognition while reducing training cost, we initialize ViT~\cite{dosovitskiy2020image} and Swin~\cite{liu2021swin} with ImageNet-1K pretrained weights~\cite{deng2009imagenet} and freeze the main backbone parameters.
Mona modules are inserted after the self-attention and feed-forward sublayers in ViT blocks, and after the window attention and MLP sublayers in Swin blocks.

To keep the overall adaptation lightweight, only task-specific components are updated, including the Mona projections, multi-scale depthwise convolution branches, the scale router, class-wise route anchors, and the classification head.
The online confusion graph is updated from model predictions during training and is not a trainable parameter.

\subsection{Routing Signature Generation}

Each Mona module contains three depthwise convolution branches with kernel sizes $3\times3$, $5\times5$, and $7\times7$.
Let $\mathcal{Q}=\{3,5,7\}$ denote the set of branch scales, and let $\mathbf{B}_q(\mathbf{h}_i^m)$ be the response of branch $q$ for sample $i$ at the $m$-th Mona module.
As shown in Fig.~\ref{fig:tldr2}(b), the original Mona statically averages multi-scale branch responses:
\begin{equation}
\mathbf{z}_{i,\mathrm{mona}}^m
=
\frac{1}{|\mathcal{Q}|}
\sum_{q\in\mathcal{Q}}
\mathbf{B}_q(\mathbf{h}_i^m).
\end{equation}

To accommodate sample-specific scale preferences under substantial intra-class appearance variations, RouteGraph-Mona replaces this static fusion with a lightweight sample-adaptive scale router, as illustrated in Fig.~\ref{fig:tldr2}(c).
For each sample, $\mathbf{p}_i^m$ is obtained by fusing a global context (the CLS token for ViT or mean-pooled tokens for Swin) with mean-pooled projected tokens.
The router then predicts:
\begin{equation}
\boldsymbol{\alpha}_i^m
=
\operatorname{softmax}
\left(
\mathbf{W}_r^m\mathbf{p}_i^m+\mathbf{b}_r^m
\right),
\end{equation}
where
$\boldsymbol{\alpha}_i^m=
[\alpha_{i,3}^m,\alpha_{i,5}^m,\alpha_{i,7}^m]$
and
$\sum_{q\in\mathcal{Q}}\alpha_{i,q}^m=1$.
The adaptive multi-scale response is computed as:
\begin{equation}
\mathbf{z}_i^m
=
\sum_{q\in\mathcal{Q}}
\alpha_{i,q}^m
\mathbf{B}_q(\mathbf{h}_i^m).
\end{equation}
When $\alpha_{i,q}^m=1/|\mathcal{Q}|$, this formulation reduces to Mona's static average fusion.

For a backbone with $L$ Transformer blocks, we collect routing weights from all $M=2L$ inserted Mona modules and compute their mean and standard deviation across modules:
\begin{equation}
\boldsymbol{\mu}_i
=
\frac{1}{M}
\sum_{m=1}^{M}
\boldsymbol{\alpha}_i^m,
\quad
\boldsymbol{\sigma}_i
=
\sqrt{
\frac{1}{M}
\sum_{m=1}^{M}
\left(
\boldsymbol{\alpha}_i^m-\boldsymbol{\mu}_i
\right)^{\odot 2}
}.
\end{equation}
The operations in $\boldsymbol{\sigma}_i$ are element-wise.
The final route signature is obtained by concatenation followed by $\ell_2$ normalization:
\begin{equation}
\mathbf{r}_i
=
\frac{
[\boldsymbol{\mu}_i;\boldsymbol{\sigma}_i]
}{
\left\|
[\boldsymbol{\mu}_i;\boldsymbol{\sigma}_i]
\right\|_2
}.
\end{equation}
Thus, $\mathbf{r}_i$ compactly characterizes sample-specific routing preferences across Mona modules.

\subsection{Class-wise Route Anchors}

For mineral images, samples from the same category may differ significantly in shape, color, and texture.
To regularize category-level routing tendencies without directly constraining visual feature embeddings, we introduce class-wise route anchors in the route space.
These anchors serve as compact route prototypes.
In the general formulation, each class $c$ is associated with $K$ learnable route anchors:
\begin{equation}
\mathcal{A}_c
=
\{\mathbf{a}_{c,k}\}_{k=1}^{K}.
\end{equation}

In our main experiments, we set $K=1$, which corresponds to one compact route prototype for each mineral category.
This setting is simple, stable, and empirically sufficient for the evaluated mineral datasets, while the formulation still allows multiple anchors when finer-grained route prototypes are required.

The route similarity between sample $i$ and class $c$ is computed using the closest route anchor:
\begin{equation}
\rho_{ic}
=
\max_{k}
\operatorname{cos}
\left(
\mathbf{r}_i,
\mathbf{a}_{c,k}
\right),
\end{equation}
where $\operatorname{cos}(\cdot,\cdot)$ denotes cosine similarity.
When $K=1$, this similarity reduces to the cosine similarity between the sample route signature and the class-level route prototype.
Let $\boldsymbol{\rho}_i=[\rho_{i1},\ldots,\rho_{iC}]$ denote the route-level class similarity vector.
To encourage samples to align with their category-level route prototype, we define the anchor loss as:
\begin{equation}
\mathcal{L}_{anchor}
=
\frac{1}{N}
\sum_{i=1}^{N}
\left[
\ell_{\mathrm{route}}(\boldsymbol{\rho}_i,y_i)
+
\left[
m_g
+
\max_{c\ne y_i}\rho_{ic}
-
\rho_{iy_i}
\right]_+
\right],
\end{equation}
where $\ell_{\mathrm{route}}$ denotes the cross-entropy loss computed on the route-level similarity vector, and $m_g$ is the route margin.

\subsection{Online Confusion-guided RouteGraph Regularization}

Some mineral categories are visually similar and are more likely to be confused by the classifier.
To focus route-level regularization on such difficult class pairs, we construct an online confusion graph $\mathbf{G}$ during training.
The graph is initialized as a row-normalized off-diagonal matrix, where each entry $\mathbf{G}_{uv}$ measures the confusion tendency from class $u$ to class $v$.

For each mini-batch, we estimate confusion using the detached prediction distribution, so that the graph reflects current model predictions without introducing extra gradient paths.
The diagonal prediction probability is removed and the remaining probabilities are normalized:
\begin{equation}
\tilde{p}_i(v)
=
\frac{
\mathbb{I}[v\ne y_i]\,\operatorname{sg}(p_i(v))
}{
\sum_{c\ne y_i}\operatorname{sg}(p_i(c))+\epsilon
},
\end{equation}
where $p_i(v)$ is the predicted probability of class $v$, $\operatorname{sg}(\cdot)$ denotes stop-gradient, and $\epsilon$ is a small constant for numerical stability.
The confusion graph is updated with momentum:
\begin{equation}
\mathbf{G}_{uv}
\leftarrow
\beta \mathbf{G}_{uv}
+
(1-\beta)
\mathbb{E}_{i:y_i=u}
[
\tilde{p}_i(v)
],
\quad u\ne v.
\end{equation}
After row-wise re-normalization, a larger $\mathbf{G}_{uv}$ indicates stronger confusion from class $u$ to class $v$.

Using this graph, we impose a confusion-weighted route margin:
\begin{equation}
\mathcal{L}_{graph}
=
\frac{1}{N}
\sum_{i=1}^{N}
\sum_{v\ne y_i}
\mathbf{G}_{y_i v}
\left[
m_g
+
\rho_{iv}
-
\rho_{iy_i}
\right]_+ .
\end{equation}
This loss assigns larger penalties to confusing category pairs and encourages stronger route-level separation.

\subsection{Training Objective and Inference Stage Details}

To combine classification learning with route-space regularization, the final training objective is:
\begin{equation}
\mathcal{L}
=
\mathcal{L}_{ce}
+
\lambda_m
\left(
\mathcal{L}_{anchor}
+
\mathcal{L}_{graph}
\right),
\end{equation}
where $\mathcal{L}_{ce}$ is the standard classification loss and $\lambda_m$ controls the strength of route-level regularization.
The anchor loss guides class-level routing tendencies, while the graph loss enlarges the separation between confusing classes.

During training, the main backbone parameters are frozen, while Mona modules, scale routers, route anchors, and the classification head are updated by back-propagation.
The online confusion graph is updated from detached model predictions and is not trainable.
During inference, no labels, graph updates, or auxiliary loss computations are required.
The adaptive scale router remains active, and the model predicts the class using the standard classification logits.
\section{Experiments}

\subsection{Datasets}

We evaluate RouteGraph-Mona on three public mineral image classification datasets: Minet~\cite{brempong2019minerals}, MinetV2~\cite{youcefattallah2022minerals}, and MineralPhotos~\cite{geillon2024mineralphotos}.
Before splitting, we identify exact duplicate images by file-hash matching and remove duplicate entries with cross-class label conflicts to reduce potential data leakage and ambiguous supervision.
The dataset sizes below correspond to the image sets used in our experiments.

\noindent\textbf{Minet} contains 955 hand-specimen images from seven categories: biotite, bornite, chrysocolla, malachite, muscovite, pyrite, and quartz.
It is small and class-imbalanced, making it suitable for evaluating limited-data adaptation.

\noindent\textbf{MinetV2} follows the same seven-class taxonomy and contains 2,091 images with higher diversity.
Compared with Minet, it shows stronger intra-class variation and inter-class ambiguity, such as chrysocolla--malachite and muscovite--quartz similarities.

\noindent\textbf{MineralPhotos} contains 39,354 images from 15 categories, providing a larger and more diverse setting for evaluating RouteGraph-Mona.

\subsection{Compared Methods}

We compare RouteGraph-Mona with two groups of baselines: general fine-tuning methods and representative mineral image classification methods.
The fine-tuning baselines include linear probing, full fine-tuning, adapter tuning \cite{houlsby2019parameter}, LoRA \cite{hu2022lora}, and Mona \cite{yin2025}.
Linear probing updates only the classification head, while full fine-tuning updates the entire pre-trained backbone.
Adapter tuning and LoRA represent widely used parameter-efficient fine-tuning strategies.
Mona is the most direct baseline, since RouteGraph-Mona is built on top of Mona by introducing adaptive routing, route-anchor regularization, and confusion-guided route-margin regularization.
For mineral image classification baselines, we include DenseNet-201 \cite{huang2017densely}, EfficientNet-B0 \cite{tan2019efficientnet}, YOLOv8-cls \cite{long2025yolov8cls}, and SwinMin \cite{swinmin2024}.
These methods cover conventional convolutional networks, efficient convolutional architectures, modern classification frameworks, and a Transformer-based model designed for mineral recognition.

\subsection{Experimental Settings}

We use ViT-B/16~\cite{dosovitskiy2020image} and Swin-B~\cite{liu2021swin} as pretrained backbones.
Each cleaned dataset is split into training, validation, and test sets at 60\%/20\%/20\%, with exact-hash grouping to prevent duplicate leakage.
All experiments are repeated with seeds 0, 666, and 2026.

Accuracy is the main evaluation metric.
For each run, the checkpoint with the lowest validation loss is evaluated once on the test set.
Models supporting pretraining use ImageNet-1K weights~\cite{deng2009imagenet}.
For fairness, all fine-tuning baselines share the same splits, input resolution, augmentation, optimizer, early-stopping criterion, seeds, and checkpoint-selection rule.
Mineral classification baselines are trained on the same cleaned splits following standard settings or official implementations when available.

Unless otherwise specified, we use $224\times224$ inputs, AdamW~\cite{loshchilov2019decoupled}, a weight decay of $1\times10^{-4}$, and an early-stopping patience of 15.
For RouteGraph-Mona, we set one route anchor per class, $\lambda_m=0.03$, $m_g=0.05$, and $\beta=0.95$.


\subsection{Main Comparison with Fine-Tuning Methods}

\newcommand{\st}[2]{#1{\scriptsize$\pm$#2}}
\newcommand{\best}[1]{\textbf{#1}}
\newcommand{\second}[1]{\underline{#1}}
\newcommand{\gain}[1]{{\scriptsize$\uparrow$#1}}

\begin{table}[t]
\centering
\caption{Comparison of fine-tuning methods on three mineral image datasets.
Accuracy is reported as mean$\pm$std over three independent runs.
The best and second-best results under each backbone are marked in \textbf{bold}
and \underline{underlined}, respectively.}
\label{tab:ft_results}
\small
\setlength{\tabcolsep}{3.6pt}
\renewcommand{\arraystretch}{1.08}
\resizebox{\textwidth}{!}{%
\begin{tabular}{@{}lcccccc@{}}
\toprule
\multirow{2}{*}{Method}
& \multicolumn{3}{c}{ViT-B/16}
& \multicolumn{3}{c}{Swin-B} \\
\cmidrule(lr){2-4}\cmidrule(l){5-7}
& Minet & MinetV2 & MineralPhotos
& Minet & MinetV2 & MineralPhotos \\
\midrule

Linear
& \st{87.04}{1.04} & \st{72.17}{1.45} & \st{63.77}{0.27}
& \st{87.88}{0.41} & \st{71.78}{1.35} & \st{64.51}{0.66} \\

Full FT
& \st{90.74}{0.95} & \st{76.73}{4.46} & \st{72.91}{0.25}
& \st{92.93}{1.24} & \best{\st{80.50}{1.06}} & \st{80.98}{0.82} \\

Adapter
& \st{90.57}{1.67} & \second{\st{76.81}{1.28}} & \second{\st{76.02}{0.73}}
& \st{91.75}{0.63} & \st{78.93}{0.89} & \st{79.63}{0.71} \\

LoRA
& \second{\st{91.25}{0.48}} & \st{75.79}{0.87} & \st{74.78}{0.67}
& \second{\st{93.10}{1.56}} & \st{77.59}{0.19} & \second{\st{81.05}{0.61}} \\

Mona
& \st{90.91}{0.71} & \st{76.34}{1.42} & \st{73.85}{0.27}
& \st{92.26}{0.86} & \st{78.22}{1.42} & \st{78.50}{0.49} \\

\midrule
\rowcolor{gray!10}
\textbf{RouteGraph-Mona}
& \best{\st{92.09}{1.05}} & \best{\st{77.12}{2.87}} & \best{\st{76.25}{0.51}}
& \best{\st{93.94}{0.71}} & \second{\st{79.64}{0.98}} & \best{\st{81.09}{0.51}} \\

\bottomrule
\end{tabular}%
}
\end{table}

As shown in Table~\ref{tab:ft_results}, RouteGraph-Mona improves the mean accuracy over its direct baseline Mona across all three mineral datasets and both visual backbones.
Under ViT-B/16, the gains over Mona are 1.18, 0.78, and 2.40 percentage points on Minet, MinetV2, and MineralPhotos, respectively; under Swin-B, the corresponding gains are 1.68, 1.42, and 2.59 percentage points.

RouteGraph-Mona also outperforms linear probing and provides stable gains over generic parameter-efficient fine-tuning methods such as adapter tuning and LoRA.
Compared with full fine-tuning, it achieves competitive performance in most settings while keeping the main pretrained backbone frozen.
These results indicate that mineral image classification benefits from explicit route-space regularization in addition to general feature adaptation.
In addition, on cross-dataset evaluation between Minet and MinetV2, RouteGraph-Mona improves Mona from 73.35\% to 73.98\% for Minet $\rightarrow$ MinetV2 and from 94.44\% to 94.61\% for MinetV2 $\rightarrow$ Minet, suggesting mild robustness under dataset shift.

\subsection{Class-Balanced Evaluation and Statistical Analysis}

Since mineral datasets are often class-imbalanced, we further report macro-F1 and mean class accuracy (mA) for the main Swin-B setting on MinetV2 using the same three seeds as Table~\ref{tab:ft_results}.
Mona obtains 78.22$\pm$1.42 accuracy, 77.36$\pm$1.36 macro-F1, and 77.38$\pm$1.53 mA, while RouteGraph-Mona obtains 79.64$\pm$0.98 accuracy, 79.00$\pm$0.84 macro-F1, and 78.97$\pm$1.12 mA.
Beyond the 1.42 percentage-point gain in overall accuracy, RouteGraph-Mona improves macro-F1 and mA by 1.64 and 1.59 percentage points, respectively.
The consistent gains across overall accuracy and both class-balanced metrics suggest that the improvement also extends to more balanced class-wise recognition performance.
A paired comparison over matched seeds shows positive gains, although the paired $t$-test does not reach statistical significance due to the small number of seeds ($p=0.151$ for accuracy).
We therefore treat these results as supporting evidence of a consistent trend rather than a standalone claim of statistical significance.

\subsection{Accuracy--Efficiency Trade-off}

\begin{figure}[t]
\centering
\includegraphics[width=0.96\linewidth]{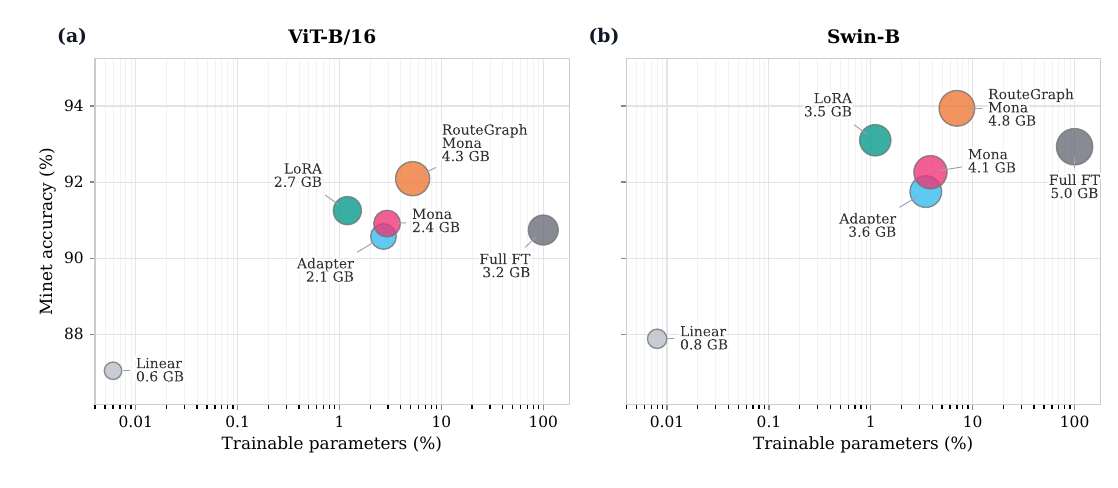}
\caption{Accuracy--parameter trade-off on Minet under two visual backbones.
The x-axis denotes the percentage of trainable parameters, the y-axis denotes the mean test accuracy over three independent runs, and the bubble size represents the peak GPU memory measured on seed 0 using the same training-mode profiling protocol.
Orange markers indicate RouteGraph-Mona.}
\label{fig:accuracy_parameter_tradeoff}
\end{figure}

Figure~\ref{fig:accuracy_parameter_tradeoff} compares accuracy, trainable parameter ratio, and peak GPU memory.
Under both ViT-B/16 and Swin-B, RouteGraph-Mona improves the accuracy over Mona while retaining a low trainable-parameter ratio.
This indicates that the performance gain does not rely on large-scale unfreezing of the pretrained backbone, but is achieved through lightweight adaptation and route-space regularization.
The bubble sizes also show that introducing the scale router and route-space regularization incurs additional training-time memory compared with Mona, while preserving the parameter-efficient update scheme.
RouteGraph-Mona updates only lightweight adaptation components, scale routers, route anchors, and the classification head, without large-scale backbone updates.
During inference, no labels, graph updates, or auxiliary losses are required; the scale router remains active, and predictions use the classification logits.

\subsection{Comparison with Mineral Image Classification Methods}

\begin{table}[t]
\centering
\caption{Comparison with representative mineral image classification methods on three mineral datasets.
Accuracy is reported as mean$\pm$std over three independent runs.
The best and second-best results are marked in \textbf{bold} and \underline{underlined}, respectively.}
\label{tab:general_results}
\small
\setlength{\tabcolsep}{5.5pt}
\renewcommand{\arraystretch}{1.08}
\begin{tabular}{@{}lccc@{}}
\toprule
Method
& Minet
& MinetV2
& MineralPhotos \\
\midrule
DenseNet-201
& 89.90$\pm$0.41
& 75.55$\pm$1.85
& 78.84$\pm$0.24 \\

EfficientNet-B0
& 90.57$\pm$2.49
& 76.10$\pm$0.95
& 76.51$\pm$1.29 \\

YOLOv8-cls
& 85.52$\pm$1.86
& 68.87$\pm$2.03
& 60.54$\pm$0.45 \\

SwinMin
& \underline{91.41$\pm$1.09}
& \underline{79.01$\pm$2.78}
& \underline{80.32$\pm$1.56} \\

\midrule
\rowcolor{gray!10}
\textbf{RouteGraph-Mona}
& \textbf{93.94$\pm$0.71}
& \textbf{79.64$\pm$0.98}
& \textbf{81.09$\pm$0.51} \\
\bottomrule
\end{tabular}
\end{table}

We further compare RouteGraph-Mona with representative mineral image classification methods, including DenseNet-201, EfficientNet-B0, YOLOv8-cls, and SwinMin.
These baselines cover conventional convolutional networks, efficient convolutional architectures, modern classification frameworks, and a Transformer-based model specifically designed for mineral image recognition.
As reported in Table~\ref{tab:general_results}, \mbox{RouteGraph-Mona} achieves the best accuracy among the compared mineral classification baselines on all three datasets.

Specifically, RouteGraph-Mona obtains 93.94\%, 79.64\%, and 81.09\% accuracy on Minet, MinetV2, and MineralPhotos, respectively.
Compared with the strongest mineral classification baseline SwinMin, RouteGraph-Mona improves the accuracy by 2.53, 0.63, and 0.77 percentage points.
This comparison shows that adapting strong pre-trained visual backbones with route-aware regularization provides a promising strategy for mineral image classification.

\subsection{Ablation Study}

\begin{table}[t]
\centering
\caption{Ablation study on Minet and MinetV2 using the Swin-B backbone.
Accuracy is reported as mean$\pm$std over three runs.}
\label{tab:ablation}
\small
\setlength{\tabcolsep}{3.6pt}
\renewcommand{\arraystretch}{1.05}
\begin{tabular}{lccc cc}
\toprule
\multirow{2}{*}{Variant}
& \multicolumn{3}{c}{Component}
& \multicolumn{2}{c}{Accuracy} \\
\cmidrule(lr){2-4}\cmidrule(l){5-6}
& Router & Anchor & Conf. Margin & Minet & MinetV2 \\
\midrule
Mona
& -- & -- & --
& 92.26$\pm$0.86 & 78.22$\pm$1.42 \\
+ Adaptive Routing
& \checkmark & -- & --
& 92.42$\pm$1.01 & 78.38$\pm$1.57 \\
+ Route Anchor
& \checkmark & \checkmark & --
& 92.42$\pm$0.87 & 78.46$\pm$1.71 \\
+ Confusion Margin
& \checkmark & -- & \checkmark
& 92.26$\pm$1.27 & 78.77$\pm$2.16 \\
\midrule
\rowcolor{gray!10}
\textbf{RouteGraph-Mona}
& \checkmark & \checkmark & \checkmark
& \textbf{93.94$\pm$0.71} & \textbf{79.64$\pm$0.98} \\
\bottomrule
\end{tabular}
\end{table}

We conduct an ablation study to examine the contribution of adaptive routing, class-wise route anchors, and the confusion-guided route margin.
As shown in Table~\ref{tab:ablation}, adding adaptive routing to Mona brings small but consistent gains on both Minet and MinetV2, indicating that replacing static multi-scale aggregation with sample-adaptive routing provides a useful basis for route-space modeling.
On top of adaptive routing, route anchors slightly improve accuracy on MinetV2 and reduce the standard deviation on Minet, suggesting that category-level route prototypes help regularize routing signatures.
The confusion-guided route margin provides a larger gain on MinetV2, where inter-class visual ambiguity is stronger, increasing the accuracy from 78.38\% to 78.77\%.

When all components are combined, RouteGraph-Mona achieves the highest accuracy among the ablation variants, reaching 93.94\% on Minet and 79.64\% on MinetV2.
The standalone gains of individual components are relatively modest, especially on the smaller Minet dataset.
This suggests that the three components play complementary roles: adaptive routing defines the route space, route anchors provide category-level references, and the confusion-guided margin emphasizes separation between visually confusing categories.

\begin{table}[t]
\centering
\caption{Feature-space versus route-space regularization on Minet and MinetV2 using Swin-B. Accuracy is reported as mean$\pm$std over three runs.}
\label{tab:route_vs_feature}
\footnotesize
\setlength{\tabcolsep}{3.0pt}
\renewcommand{\arraystretch}{1.08}
\begin{tabular}{@{}l l c c c c@{}}
\toprule
\multirow{2}{*}{Method}
& \multirow{2}{*}{Space}
& \multicolumn{2}{c}{Minet}
& \multicolumn{2}{c}{MinetV2} \\
\cmidrule(lr){3-4}\cmidrule(lr){5-6}
& & Acc. & Gain & Acc. & Gain \\
\midrule
Mona
& -- 
& 92.26$\pm$0.86 & -- 
& 78.22$\pm$1.42 & -- \\

Center Loss
& Feature
& 92.09$\pm$1.27 & $-0.17$
& 77.99$\pm$0.72 & $-0.23$ \\

SupCon
& Feature
& 91.92$\pm$2.02 & $-0.34$
& 78.93$\pm$0.49 & $+0.71$ \\

Feature Anchor+Graph
& Feature
& 91.08$\pm$2.28 & $-1.18$
& 77.44$\pm$1.80 & $-0.78$ \\

\rowcolor{gray!8}
\textbf{RouteGraph-Mona}
& \textbf{Route}
& \textbf{93.94$\pm$0.71} & \textbf{$+1.68$}
& \textbf{79.64$\pm$0.98} & \textbf{$+1.42$} \\
\bottomrule
\end{tabular}
\end{table}

\subsection{Feature-space versus Route-space Regularization}

Table~\ref{tab:route_vs_feature} compares RouteGraph-Mona with several feature-space regularization variants using the Swin-B backbone.
Center Loss~\cite{wen2016discriminative} and supervised contrastive learning~\cite{khosla2020supervised} are included as generic feature-level discriminative losses.
Center Loss slightly decreases accuracy on both datasets, while supervised contrastive learning improves MinetV2 but decreases Minet, indicating that generic feature-space constraints do not consistently benefit mineral image classification.

To isolate the effect of the regularization space, we construct Feature Anchor+Graph, which uses the same anchor-based prototype constraint, confusion-weighted margin, and online graph construction as RouteGraph-Mona, but applies them to visual feature embeddings rather than routing signatures.
This provides a matched comparison between feature-space and route-space regularization.

As shown in Table~\ref{tab:route_vs_feature}, Feature Anchor+Graph performs worse than RouteGraph-Mona on both datasets, indicating that the gains do not simply arise from prototype- or graph-based regularization, but from applying these structural constraints in the route space induced by adaptive multi-scale aggregation.

\begin{figure}[t]
\centering
\includegraphics[width=0.98\linewidth]{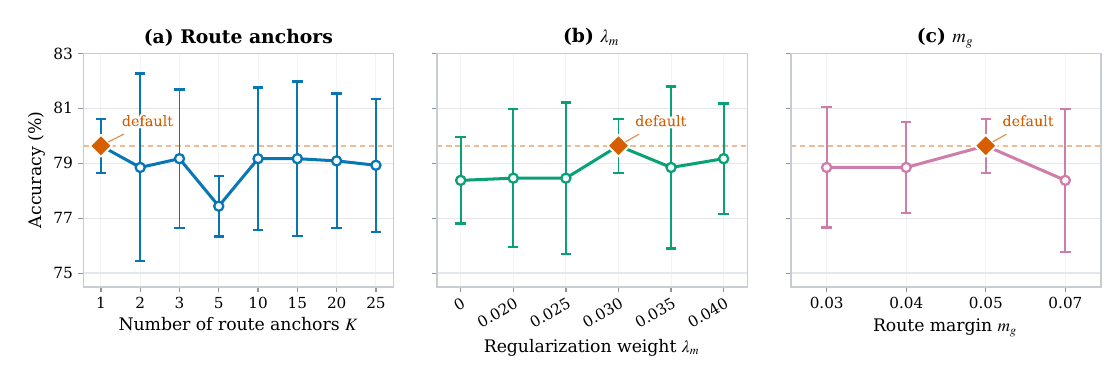}
\caption{Hyperparameter sensitivity on MinetV2 with the Swin-B backbone.
Accuracy is reported as mean$\pm$std over three runs, and the diamond marker indicates the default setting.}
\label{fig:hyperparameter_sensitivity}
\end{figure}

\begin{figure}[t]
\centering
\includegraphics[width=0.95\linewidth]{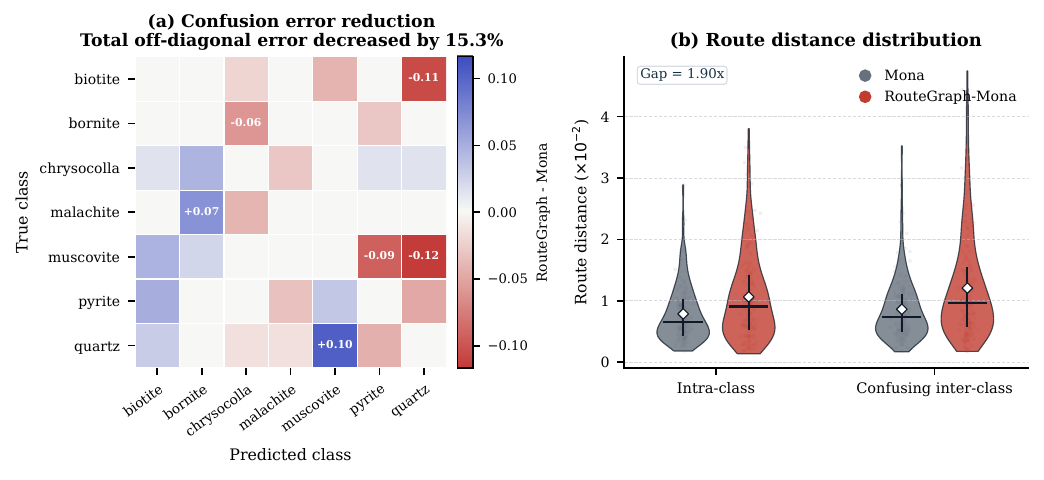}
\caption{Visualization analysis on MinetV2 using the Swin-B backbone.
(a) Confusion change from Mona to RouteGraph-Mona.
Negative values indicate reduced off-diagonal confusion errors, and the total off-diagonal error is decreased by 15.3\%.
(b) Route-distance distributions in the routing space.
RouteGraph-Mona widens the separation gap between intra-class and confusing inter-class distances.}
\label{fig:visualization_analysis}
\end{figure}

\subsection{Effect of Confusion Graph Construction}

\begin{table}[t]
\centering
\caption{Effect of different graph construction strategies on MinetV2 using the Swin-B backbone. Accuracy is reported as mean$\pm$std over three independent runs. The online confusion graph corresponds to the main RouteGraph-Mona setting.}
\label{tab:graph_construction}
\small
\setlength{\tabcolsep}{6pt}
\renewcommand{\arraystretch}{1.08}
\begin{tabular}{lc}
\toprule
Graph Strategy & Accuracy \\
\midrule
No Graph & 78.46$\pm$1.71 \\
Uniform Graph & 78.30$\pm$2.86 \\
Random Graph & 78.69$\pm$2.61 \\
\rowcolor{gray!10}
\textbf{Online Confusion Graph}
& \textbf{79.64$\pm$0.98} \\
\bottomrule
\end{tabular}
\end{table}

We compare different graph construction strategies on MinetV2 with the Swin-B backbone, where stronger inter-class visual ambiguity makes confusion-aware regularization more informative.
As shown in Table~\ref{tab:graph_construction}, the no-graph setting achieves 78.46$\pm$1.71\% accuracy, while the uniform graph slightly decreases the result to 78.30$\pm$2.86\%, indicating that assigning equal weights to all negative classes does not provide a stable benefit.
Random graph weighting increases the mean accuracy to 78.69\%, but also raises the standard deviation to 2.61 and lacks semantic correspondence to actual category confusion.

In contrast, the online confusion graph achieves the best result of 79.64$\pm$0.98\%, improving the no-graph setting by 1.18 percentage points while exhibiting the lowest variance among the evaluated graph strategies.
These results suggest that prediction-derived confusion provides a more effective and stable weighting strategy than fixed uniform or random graph weighting.
By upweighting route-margin penalties for confusing category pairs, the online graph focuses regularization on mineral categories that are more likely to be visually confused.

\subsection{Hyperparameter Sensitivity}

Figure~\ref{fig:hyperparameter_sensitivity} analyzes the effects of the number of route anchors $K$, the regularization weight $\lambda_m$, and the route margin $m_g$ on model performance.
A single route anchor per class already provides strong mean accuracy, while increasing $K$ does not produce a consistent improvement; in particular, the performance drops noticeably at $K=5$.
For the regularization weight, the default setting $\lambda_m=0.03$ achieves competitive performance, while the other evaluated values show only moderate fluctuations.
For the route margin, $m_g=0.05$ performs well, whereas further increasing it to $0.07$ leads to a performance decrease.
Overall, the default settings provide stable performance, and increasing the number of anchors or the regularization strength does not consistently yield additional gains.
Given the overlap among the error bars across several settings, these results are interpreted mainly as overall sensitivity trends rather than statistically significant differences.

\subsection{Analysis of Route-Space Geometry}

We further examine whether RouteGraph-Mona improves the route-space geometry behind classification decisions.
Figure~\ref{fig:visualization_analysis} analyzes RouteGraph-Mona from the perspectives of class confusion and route-space geometry.

As shown in Figure~\ref{fig:visualization_analysis}, RouteGraph-Mona reduces the total off-diagonal confusion error by 15.3\% compared with Mona, indicating that the proposed method alleviates misclassification among visually similar mineral categories.
Quantitatively, RouteGraph-Mona increases the confusing inter-class route distance from $0.862\times10^{-2}$ to $1.206\times10^{-2}$ and widens the separation gap from $0.076\times10^{-2}$ to $0.144\times10^{-2}$, making the route-space separation gap $1.90\times$ as large as that of Mona.
This suggests that the improvement mainly comes from better separation of visually confusing categories in the route space, rather than simply compressing samples from the same class.

\section{Conclusion}

In this paper, we propose RouteGraph-Mona, a lightweight route-space regularization framework for mineral image classification. The proposed method replaces Mona's static multi-scale aggregation with sample-adaptive routing and applies class-wise route anchor constraints and an online confusion-guided route margin to routing signatures formed by routing weights. By regularizing routing signatures rather than visual feature embeddings directly, RouteGraph-Mona improves the relative route-space separation between visually confusing mineral categories. Experiments on three mineral image datasets and two visual backbones show that RouteGraph-Mona improves the mean accuracy over Mona across the evaluated settings and achieves competitive performance compared with representative fine-tuning and mineral image classification baselines. In our experiments, one route anchor per class is empirically sufficient; nevertheless, future work may explore class-adaptive anchor allocation and more robust confusion graph updating strategies on larger and more diverse mineral datasets.

\bibliographystyle{splncs04}
\bibliography{bibliography}

@article{lou2020review,
author={Lou, Wei and Zhang, Dexian and Bayless, Richard C},
title={Review of mineral recognition and its future},
journal={Appl. Geochem.},
volume={122},
pages={104727},
year={2020}
}

@article{zeng2021mineral,
author={Zeng, Xiang and Xiao, Yancong and Ji, Xiaohui and Wang, Gongwen},
title={Mineral identification based on deep learning that combines image and mohs hardness},
journal={Minerals},
volume={11},
pages={506},
year={2021}
}

@article{jia2021mineralphotos,
author={Jia, Liqin and Yang, Mei and Meng, Fang and He, Mingyue and Liu, Hongmin},
title={Mineral photos recognition based on feature fusion and online hard sample mining},
journal={Minerals},
volume={11},
pages={1354},
year={2021}
}

@article{wu2022multilabel,
author={Wu, Baokun and Ji, Xiaohui and He, Mingyue and Yang, Mei and Zhang, Zhaochong and Chen, Yan and Wang, Yuzhu and Zheng, Xinqi},
title={Mineral identification based on multi-label image classification},
journal={Minerals},
volume={12},
pages={1338},
year={2022}
}

@article{asiedu2021minet,
  title={MiNet: a convolutional neural network for identifying and categorising minerals},
  author={Brempong, Emmanuel Asiedu and Agangiba, Millicent and Aikins, Daniel},
  journal={arXiv preprint arXiv:2111.11260},
  year={2021}
}

@article{swinmin2024,
author={Jia, Liqin and Chen, Feng and Yang, Mei and Meng, Fang and He, Mingyue and Liu, Hongmin},
title={SwinMin: {A} mineral recognition model incorporating convolution and multi-scale contexts into Swin Transformer},
journal={Comput. Geosci.},
volume={184},
pages={105532},
year={2024}
}

@inproceedings{yosinski2014transferable,
author={Yosinski, Jason and Clune, Jeff and Bengio, Yoshua and Lipson, Hod},
title={How transferable are features in deep neural networks?},
booktitle={NeurIPS},
pages={3320--3328},
year={2014}
}

@inproceedings{kornblith2019better,
author={Kornblith, Simon and Shlens, Jonathon and Le, Quoc V.},
title={Do better ImageNet models transfer better?},
booktitle={CVPR},
pages={2661--2671},
year={2019}
}

@inproceedings{kolesnikov2020big,
author={Kolesnikov, Alexander and Beyer, Lucas and Zhai, Xiaohua and Puigcerver, Joan and Yung, Jessica and Gelly, Sylvain and Houlsby, Neil},
title={Big Transfer ({BiT}): General visual representation learning},
booktitle={ECCV},
pages={491--507},
year={2020}
}

@inproceedings{yin2025,
  title={5\%> 100\%: Breaking performance shackles of full fine-tuning on visual recognition tasks},
  author={Yin, Dongshuo and Hu, Leiyi and Li, Bin and Zhang, Youqun and Yang, Xue},
  booktitle={Proceedings of the Computer Vision and Pattern Recognition Conference},
  pages={20071--20081},
  year={2025}
}

@inproceedings{houlsby2019parameter,
author={Houlsby, Neil and Giurgiu, Andrei and Jastrzebski, Stanislaw and Morrone, Bruna and De Laroussilhe, Quentin and Gesmundo, Andrea and Attariyan, Mona and Gelly, Sylvain},
title={Parameter-efficient transfer learning for NLP},
booktitle={ICML},
pages={2790--2799},
year={2019}
}

@inproceedings{hu2022lora,
author={Hu, Edward J. and Shen, Yelong and Wallis, Phillip and Allen-Zhu, Zeyuan and Li, Yuanzhi and Wang, Shean and Wang, Liang and Chen, Weizhu},
title={LoRA: Low-rank adaptation of large language models},
booktitle={ICLR},
year={2022}
}

@inproceedings{jia2022visual,
author={Jia, Menglin and Tang, Luming and Chen, Bor-Chun and Cardie, Claire and Belongie, Serge and Hariharan, Bharath and Lim, Ser-Nam},
title={Visual prompt tuning},
booktitle={ECCV},
pages={709--727},
year={2022}
}

@inproceedings{he2023parameter,
author={He, Xuehai and Li, Chunyuan and Zhang, Pengchuan and Yang, Jianwei and Wang, Xin Eric},
title={Parameter-efficient model adaptation for vision transformers},
booktitle={AAAI},
pages={817--825},
year={2023}
}

@article{long2025yolov8cls,
author={Long, Bofan and Zhou, Chunyang},
title={Enhanced mineral image classification using YOLOv8-CLS with optimized feature extraction and dataset augmentation},
journal={Informatica},
volume={49},
pages={},
year={2025}
}

@article{attallah2025mineral,
author={Attallah, Youcef and Zigh, Ehlem and Ka{\c{c}}maz, Seydi and Maazouzi, Amine},
title={Mineral classification using Dense-Net architectures: leveraging deep feature extraction for enhanced accuracy},
journal={Int. Adv. Res. Eng. J.},
volume={9},
pages={201--210},
year={2025}
}

@article{haskin1997raman,
author={Haskin, Larry A. and Wang, Alian and Rockow, Kaylynn M. and Jolliff, Bradley L. and Korotev, Randy L. and Viskupic, Karen M.},
title={Raman spectroscopy for mineral identification and quantification for in situ planetary surface analysis: A point count method},
journal={J. Geophys. Res. Planets},
volume={102},
pages={19293--19306},
year={1997}
}

@article{ali2022xrd,
author={Ali, Asif and Chiang, Yi Wai and Santos, Rafael M.},
title={{X}-ray diffraction techniques for mineral characterization: A review for engineers of the fundamentals, applications, and research directions},
journal={Minerals},
volume={12},
pages={205},
year={2022}
}

@article{ali2023sem,
author={Ali, Asif and Zhang, Ning and Santos, Rafael M.},
title={Mineral characterization using scanning electron microscopy ({SEM}): A review of the fundamentals, advancements, and research directions},
journal={Appl. Sci.},
volume={13},
pages={12600},
year={2023}
}

@article{dosovitskiy2020image,
  title={An image is worth 16x16 words: Transformers for image recognition at scale},
  author={Dosovitskiy, Alexey and Beyer, Lucas and Kolesnikov, Alexander and Weissenborn, Dirk and Zhai, Xiaohua and Unterthiner, Thomas and Dehghani, Mostafa and Minderer, Matthias and Heigold, Georg and Gelly, Sylvain and others},
  journal={arXiv preprint arXiv:2010.11929},
  year={2020}
}

@inproceedings{liu2021swin,
  title={Swin transformer: Hierarchical vision transformer using shifted windows},
  author={Liu, Ze and Lin, Yutong and Cao, Yue and Hu, Han and Wei, Yixuan and Zhang, Zheng and Lin, Stephen and Guo, Baining},
  booktitle={Proceedings of the IEEE/CVF international conference on computer vision},
  pages={10012--10022},
  year={2021}
}

@inproceedings{deng2009imagenet,
  title={Imagenet: A large-scale hierarchical image database},
  author={Deng, Jia and Dong, Wei and Socher, Richard and Li, Li-Jia and Li, Kai and Fei-Fei, Li},
  booktitle={2009 IEEE conference on computer vision and pattern recognition},
  pages={248--255},
  year={2009},
  organization={Ieee}
}

@inproceedings{huang2017densely,
author={Huang, Gao and Liu, Zhuang and van der Maaten, Laurens and Weinberger, Kilian Q.},
title={Densely connected convolutional networks},
booktitle={CVPR},
pages={4700--4708},
year={2017}
}

@inproceedings{tan2019efficientnet,
author={Tan, Mingxing and Le, Quoc V.},
title={{EfficientNet}: Rethinking model scaling for convolutional neural networks},
booktitle={ICML},
pages={6105--6114},
year={2019}
}

@inproceedings{loshchilov2019decoupled,
author={Loshchilov, Ilya and Hutter, Frank},
title={Decoupled weight decay regularization},
booktitle={ICLR},
year={2019}
}

@inproceedings{wen2016discriminative,
author={Wen, Yandong and Zhang, Kaipeng and Li, Zhifeng and Qiao, Yu},
title={A discriminative feature learning approach for deep face recognition},
booktitle={ECCV},
pages={499--515},
year={2016}
}

@inproceedings{khosla2020supervised,
author={Khosla, Prannay and Teterwak, Piotr and Wang, Chen and Sarna, Aaron and Tian, Yonglong and Isola, Phillip and Maschinot, Aaron and Liu, Ce and Krishnan, Dilip},
title={Supervised contrastive learning},
booktitle={NeurIPS},
pages={18661--18673},
year={2020}
}

@misc{brempong2019minerals,
author={Brempong, Kofi Asiedu},
title={Minerals Identification Dataset},
year={2019},
howpublished={Kaggle dataset},
note={Available at: https://www.kaggle.com/datasets/asiedubrempong/minerals-identification-dataset}
}

@misc{youcefattallah2022minerals,
author={{Youcefattallah97}},
title={Minerals Identification \& Classification},
year={2022},
howpublished={Kaggle dataset},
note={Available at: https://www.kaggle.com/datasets/youcefattallah97/minerals-identification-classification}
}

@misc{geillon2024mineralphotos,
author={Geillon, Florian},
title={Mineral Photos},
year={2024},
howpublished={Kaggle dataset},
note={Available at: https://www.kaggle.com/datasets/floriangeillon/mineral-photos}
}

\end{document}